# Learning to See in the Dark

Chen Chen  Qifeng Chen  Jia Xu  Vladlen Koltun
UIUC  Intel Labs  Intel Labs  Intel Labs

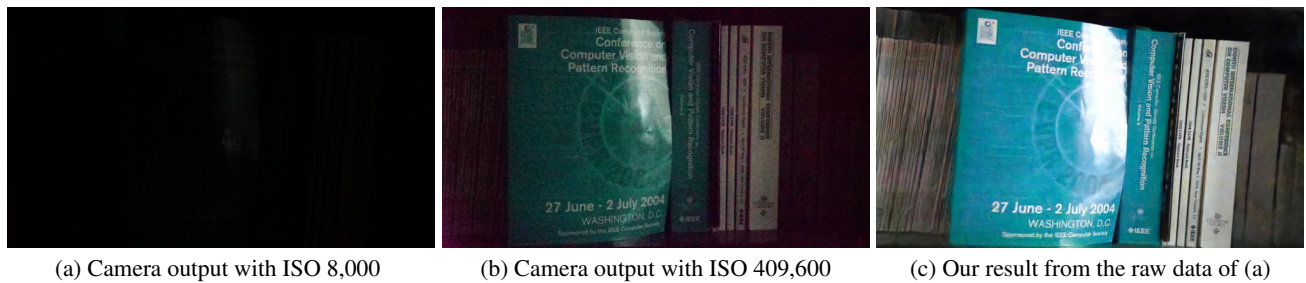

(a) Camera output with ISO 8,000  (b) Camera output with ISO 409,600  (c) Our result from the raw data of (a)

Figure 1. Extreme low-light imaging with a convolutional network. Dark indoor environment. The illuminance at the camera is $< 0.1$ lux. The Sony $\alpha$7S II sensor is exposed for 1/30 second. (a) Image produced by the camera with ISO 8,000. (b) Image produced by the camera with ISO 409,600. The image suffers from noise and color bias. (c) Image produced by our convolutional network applied to the raw sensor data from (a).

## Abstract

*Imaging in low light is challenging due to low photon count and low SNR. Short-exposure images suffer from noise, while long exposure can induce blur and is often impractical. A variety of denoising, deblurring, and enhancement techniques have been proposed, but their effectiveness is limited in extreme conditions, such as video-rate imaging at night. To support the development of learning-based pipelines for low-light image processing, we introduce a dataset of raw short-exposure low-light images, with corresponding long-exposure reference images. Using the presented dataset, we develop a pipeline for processing low-light images, based on end-to-end training of a fully-convolutional network. The network operates directly on raw sensor data and replaces much of the traditional image processing pipeline, which tends to perform poorly on such data. We report promising results on the new dataset, analyze factors that affect performance, and highlight opportunities for future work.*

## 1. Introduction

Noise is present in any imaging system, but it makes imaging particularly challenging in low light. High ISO can be used to increase brightness, but it also amplifies noise. Postprocessing, such as scaling or histogram stretching, can be applied, but this does not resolve the low signal-to-noise ratio (SNR) due to low photon counts. There are physical means to increase SNR in low light, including opening the aperture, extending exposure time, and using flash. But each of these has its own characteristic drawbacks. For example, increasing exposure time can introduce blur due to camera shake or object motion.

The challenge of fast imaging in low light is well-known in the computational photography community, but remains open. Researchers have proposed techniques for denoising, deblurring, and enhancement of low-light images [34, 16, 42]. These techniques generally assume that images are captured in somewhat dim environments with moderate levels of noise. In contrast, we are interested in extreme low-light imaging with severely limited illumination (e.g., moonlight) and short exposure (ideally at video rate). In this regime, the traditional camera processing pipeline breaks down and the image has to be reconstructed from the raw sensor data.

Figure 1 illustrates our setting. The environment is extremely dark: less than 0.1 lux of illumination at the camera. The exposure time is set to 1/30 second. The aperture is f/5.6. At ISO 8,000, which is generally considered high, the camera produces an image that is essentially black, despite the high light sensitivity of the full-frame Sony sensor. At ISO 409,600, which is far beyond the reach of most cameras, the content of the scene is discernible, but the image is dim, noisy, and the colors are distorted. As we will show, even state-of-the-art denoising techniques [32] fail to remove such noise and do not address the color bias. An alternative approach is to use a burst of images [24, 14], but burst alignment algorithms may fail in extreme low-light conditions and burst pipelines are not designed for video capture (e.g., due to the use of 'lucky imaging' within the burst).

We propose a new image processing pipeline that addresses the challenges of extreme low-light photography via a data-driven approach. Specifically, we train deep neural networks to learn the image processing pipeline for low-light raw data, including color transformations, demosaicing, noise reduction, and image enhancement. The pipeline is trained end-to-end to avoid the noise amplification and error accumulation that characterize traditional camera processing pipelines in this regime.

Most existing methods for processing low-light images were evaluated on synthetic data or on real low-light images without ground truth. To the best of our knowledge, there is no public dataset for training and testing techniques for processing fast low-light images with diverse real-world data and ground truth. Therefore, we have collected a new dataset of raw images captured with fast exposure in low-light conditions. Each low-light image has a corresponding long-exposure high-quality reference image. We demonstrate promising results on the new dataset: low-light images are amplified by up to 300 times with successful noise reduction and correct color transformation. We systematically analyze key elements of the pipeline and discuss directions for future research.

## 2. Related Work

Computational processing of low-light images has been extensively studied in the literature. We provide a short review of existing methods.

**Image denoising.** Image denoising is a well-developed topic in low-level vision. Many approaches have been proposed, using techniques such as total variation [36], wavelet-domain processing [33], sparse coding [9, 28], nuclear norm minimization [12], and 3D transform-domain filtering (BM3D) [7]. These methods are often based on specific image priors such as smoothness, sparsity, low rank, or self-similarity. Researchers have also explored the application of deep networks to denoising, including stacked sparse denoising auto-encoders (SSDA) [39, 1], trainable nonlinear reaction diffusion (TNRD) [6], multi-layer perceptrons [3], deep autoencoders [26], and convolutional networks [17, 41]. When trained on certain noise levels, these data-driven methods can compete with state-of-the-art classic techniques such as BM3D and sparse coding. Unfortunately, most existing methods have been evaluated on synthetic data, such as images with added Gaussian or salt&pepper noise. A careful recent evaluation with real data found that BM3D outperforms more recent techniques on real images [32]. Joint denoising and demosaicing has also been studied, including recent work that uses deep networks [15, 10], but these methods have been evaluated on synthetic Bayer patterns and synthetic noise, rather than real images collected in extreme low-light conditions.

In addition to single-image denoising, multiple-image denoising has also been considered and can achieve better results since more information is collected from the scene [31, 23, 19, 24, 14, 29]. In particular, Liu et al. [24] and Hasinoff et al. [14] propose to denoise a burst of images from the same scene. While often effective, these pipelines can be elaborate, involving reference image selection ('lucky imaging') and dense correspondence estimation across images. We focus on a complementary line of investigation and study how far single-image processing can be pushed.

**Low-light image enhancement.** A variety of techniques have been applied to enhance the contrast of low-light images. One classic choice is histogram equalization, which balances the histogram of the entire image. Another widely used technique is gamma correction, which increases the brightness of dark regions while compressing bright pixels. More advanced methods perform more global analysis and processing, using for example the inverse dark channel prior [8, 29], the wavelet transform [27], the Retinex model [30], and illumination map estimation [13]. However, these methods generally assume that the images already contain a good representation of the scene content. They do not explicitly model image noise and typically apply off-the-shelf denoising as a postprocess. In contrast, we consider extreme low-light imaging, with severe noise and color distortion that is beyond the operating conditions of existing enhancement pipelines.

**Noisy image datasets.** Although there are many studies of image denoising, most existing methods are evaluated on synthetic data, such as clean images with added Gaussian or salt&pepper noise. The RENOIR dataset [2] was proposed to benchmark denoising with real noisy images. However, as reported in the literature [32], image pairs in the RENOIR dataset exhibit spatial misalignment. Bursts of images have been used to reduce noise in low-light conditions [24], but the associated datasets do not contain reliable ground-truth data. The Google HDR+ dataset [14] does not target extreme low-light imaging: most images in the dataset were captured during the day. The recent Darmstadt Noise Dataset (DND) [32] aims to address the need for real data in the denoising community, but the images were captured during the day and are not suitable for evaluation of low-light image processing. To the best of our knowledge, there is no public dataset with raw low-light images and corresponding ground truth. We therefore collect such a dataset to support systematic reproducible research in this area.

| Sony α7S II | Filter array | Exposure time (s) | # images |
|---|---|---|---|
| x300 | Bayer | 1/10, 1/30 | 1190 |
| x250 | Bayer | 1/25 | 699 |
| x100 | Bayer | 1/10 | 808 |
| Fujifilm X-T2 | Filter array | Exposure time (s) | # images |
| x300 | X-Trans | 1/30 | 630 |
| x250 | X-Trans | 1/25 | 650 |
| x100 | X-Trans | 1/10 | 1117 |

Table 1. The See-in-the-Dark (SID) dataset contains 5094 raw short-exposure images, each with a reference long-exposure image. The images were collected by two cameras (top and bottom). From left to right: ratio of exposure times between input and reference images, filter array, exposure time of input image, and number of images in each condition.

## 3. See-in-the-Dark Dataset

We collected a new dataset for training and benchmarking single-image processing of raw low-light images. The See-in-the-Dark (SID) dataset contains 5094 raw short-exposure images, each with a corresponding long-exposure reference image. Note that multiple short-exposure images can correspond to the same long-exposure reference image. For example, we collected sequences of short-exposure images to evaluate burst denoising methods. Each image in the sequence is counted as a distinct low-light image, since each such image contains real imaging artifacts and is useful for training and testing. The number of distinct long-exposure reference images in SID is 424.

The dataset contains both indoor and outdoor images. The outdoor images were generally captured at night, under moonlight or street lighting. The illuminance at the camera in the outdoor scenes is generally between 0.2 lux and 5 lux. The indoor images are even darker. They were captured in closed rooms with regular lights turned off and with faint indirect illumination set up for this purpose. The illuminance at the camera in the indoor scenes is generally between 0.03 lux and 0.3 lux.

The exposure for the input images was set between 1/30 and 1/10 seconds. The corresponding reference (ground truth) images were captured with 100 to 300 times longer exposure: i.e., 10 to 30 seconds. Since exposure times for the reference images are necessarily long, all the scenes in the dataset are static. The dataset is summarized in Table 1. A small sample of reference images is shown in Figure 2. Approximately 20% of the images in each condition are randomly selected to form the test set, and another 10% are selected for the validation set.

Images were captured using two cameras: Sony α7S II and Fujifilm X-T2. These cameras have different sensors: the Sony camera has a full-frame Bayer sensor and the Fuji camera has an APS-C X-Trans sensor. This supports evaluation of low-light image processing pipelines on images produced by different filter arrays. The resolution is 4240×2832 for Sony and 6000×4000 for the Fuji images. The Sony set was collected using two different lenses.

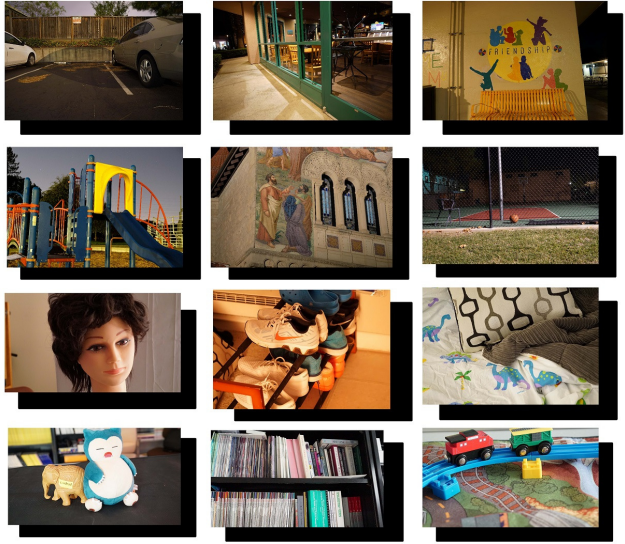

Figure 2. Example images in the SID dataset. Outdoor images in the top two rows, indoor images in the bottom rows. Long-exposure reference (ground truth) images are shown in front. Short-exposure input images (essentially black) are shown in the back. The illuminance at the camera is generally between 0.2 and 5 lux outdoors and between 0.03 and 0.3 lux indoors.

The cameras were mounted on sturdy tripods. We used mirrorless cameras to avoid vibration due to mirror flapping. In each scene, camera settings such as aperture, ISO, focus, and focal length were adjusted to maximize the quality of the reference (long-exposure) images. After a long-exposure reference image was taken, a remote smartphone app was used to decrease the exposure time by a factor of 100 to 300 for a sequence of short-exposure images. The camera was not touched between the long-exposure and the short-exposure images. We collected sequences of short-exposure images to support comparison with an idealized burst-imaging pipeline that benefits from perfect alignment.

The long-exposure reference images may still contain some noise, but the perceptual quality is sufficiently high for these images to serve as ground truth. We target applications that aim to produce perceptually good images in low-light conditions, rather than exhaustively removing all noise or maximizing image contrast.

## 4. Method

### 4.1. Pipeline

After getting the raw data from an imaging sensor, the traditional image processing pipeline applies a sequence of modules such as white balance, demosaicing, denoising, sharpening, color space conversion, gamma correction, and others. These modules are often tuned for specific cameras. Jiang et al. [18] proposed to use a large collection of local, linear, and learned (L3) filters to approximate the complex nonlinear pipelines found in modern consumer imaging systems. Yet neither the traditional pipeline nor the L3 pipeline successfully deal with fast low-light imaging, as they are not able to handle the extremely low SNR. Hasinoff et al. [14] described a burst imaging pipeline for smartphone cameras. This method can produce good results by aligning and blending multiple images, but introduces a certain level of complexity, for example due to the need for dense correspondence estimation, and may not easily extend to video capture, for example due to the use of lucky imaging.

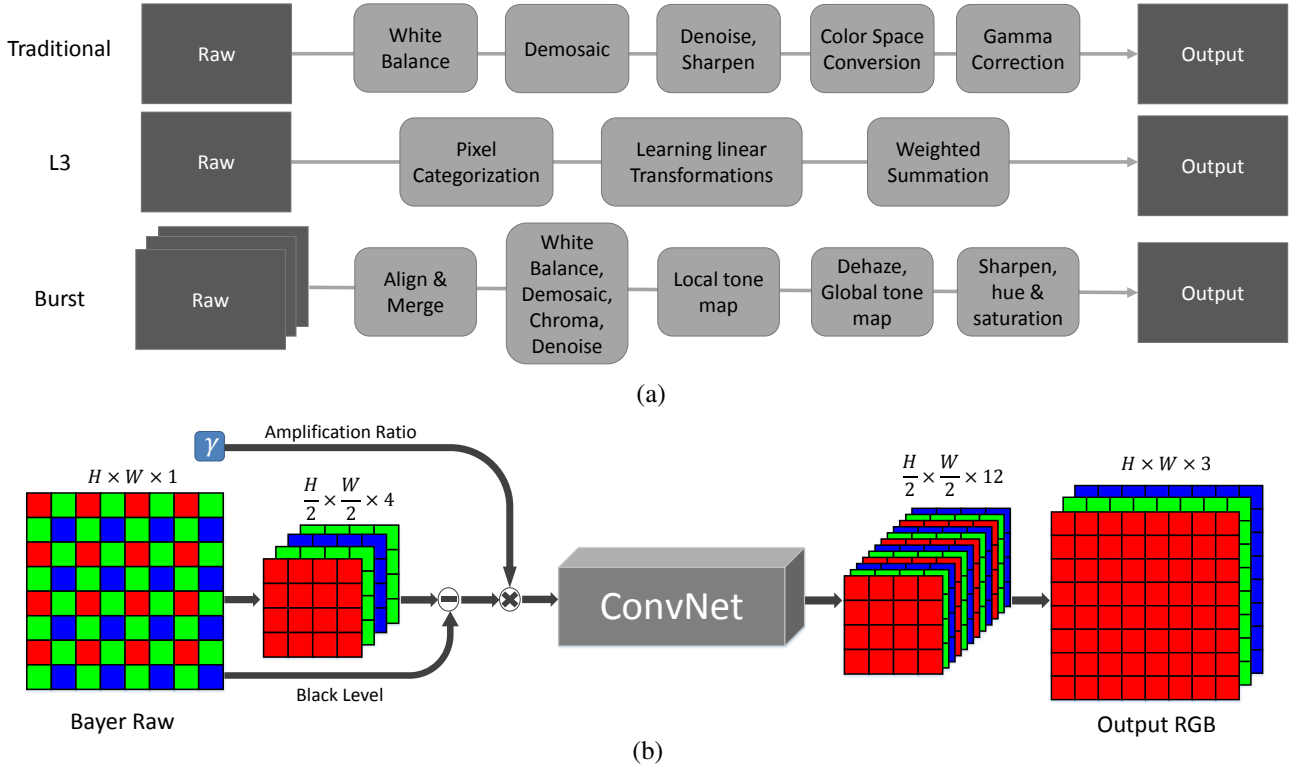

Figure 3. The structure of different image processing pipelines. (a) From top to bottom: a traditional image processing pipeline, the L3 pipeline [18], and a burst imaging pipeline [14]. (b) Our pipeline.

We propose to use end-to-end learning for direct single-image processing of fast low-light images. Specifically, we train a fully-convolutional network (FCN) [22, 25] to perform the entire image processing pipeline. Recent work has shown that pure FCNs can effectively represent many image processing algorithms [40, 5]. We are inspired by this work and investigate the application of this approach to extreme low-light imaging. Rather than operating on normal sRGB images produced by traditional camera processing pipelines, we operate on raw sensor data.

Figure 3(b) illustrates the structure of the presented pipeline. For Bayer arrays, we pack the input into four channels and correspondingly reduce the spatial resolution by a factor of two in each dimension. For X-Trans arrays (not shown in the figure), the raw data is arranged in 6×6 blocks; we pack it into 9 channels instead of 36 channels by exchanging adjacent elements. We subtract the black level and scale the data by the desired amplification ratio (e.g., x100 or x300). The packed and amplified data is fed into a fully-convolutional network. The output is a 12-channel image with half the spatial resolution. This half-sized output is processed by a sub-pixel layer to recover the original resolution [37].

After preliminary exploration, we have focused on two general structures for the fully-convolutional network that forms the core of our pipeline: a multi-scale context aggregation network (CAN) recently used for fast image processing [5] and a U-net [35]. Other work has explored residual connections [20, 34, 41], but we did not find these beneficial in our setting, possibly because our input and output are represented in different color spaces. Another consideration that affected our choice of architectures is memory consumption: we have chosen architectures that can process a full-resolution image (e.g., at 4240×2832 or 6000×4000 resolution) in GPU memory. We have therefore avoided fully-connected layers that require processing small image patches and reassembling them [26]. Our default architecture is the U-net [35].

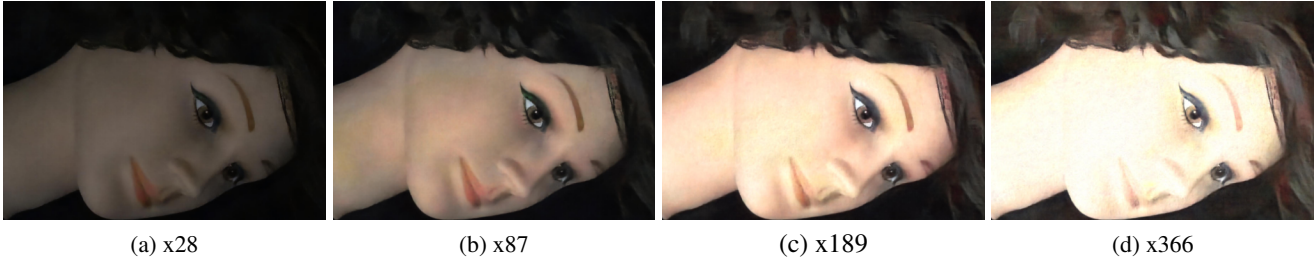

| (a) x28 | (b) x87 | (c) x189 | (d) x366 |

Figure 4. The effect of the amplification factor on a patch from an indoor image in the SID dataset (Sony x100 subset). The amplification factor is provided as an external input to our pipeline, akin to the ISO setting in cameras. Higher amplification factors yield brighter images. This figure shows the output of our pipeline with different amplification factors.

The amplification ratio determines the brightness of the output. In our pipeline, the amplification ratio is set externally and is provided as input to the pipeline, akin to the ISO setting in cameras. Figure 4 shows the effect of different amplification ratios. The user can adjust the brightness of the output image by setting different amplification factors. At test time, the pipeline performs blind noise suppression and color transformation. The network outputs the processed image directly in sRGB space.

### 4.2. Training

We train the networks from scratch using the $L_1$ loss and the Adam optimizer [21]. During training, the input to the network is the raw data of the short-exposed image and the ground truth is the corresponding long-exposure image in sRGB space (processed by `libraw`, a raw image processing library). We train one network for each camera. The amplification ratio is set to be the exposure difference between the input and reference images (e.g., x100, x250, or x300) for both training and testing. In each iteration, we randomly crop a $512 \times 512$ patch for training and apply random flipping and rotation for data augmentation. The learning rate is initially set to $10^{-4}$ and is reduced to $10^{-5}$ after 2000 epochs. Training proceeds for 4000 epochs.

## 5. Experiments

### 5.1. Qualitative results and perceptual experiments

**Comparison to traditional pipeline.** Our initial baseline is the traditional camera processing pipeline, with amplification prior to quantization. (We use the same amplification ratio as the one given to our pipeline.) Qualitative comparisons to this baseline are shown in Figures 5, 6, and 7. Images produced by the traditional pipeline in extreme low-light conditions suffer from severe noise and color distortion.

**Comparison to denoising and burst processing.** The natural next step is to apply an existing denoising algorithm post-hoc to the output of the traditional pipeline. A careful recent evaluation on real data has shown that BM3D [7] outperforms more recent denoising models on real images [32]. We thus use BM3D as the reference denoising algorithm. Figure 7 illustrates the results. Note that BM3D is a non-blind denoising method and requires the noise level to be specified extrinsically as a parameter. A small noise level setting may leave perceptually significant noise in the image, while a large level may over-smooth. As shown in Figure 7, the two effects can coexist in the same image, since uniform additive noise is not an appropriate model for real low-light images. In contrast, our pipeline performs blind noise suppression that can locally adapt to the data. Furthermore, post-hoc denoising does not address other artifacts present in the output of the traditional pipeline, such as color distortion.

We also compare to burst denoising [24, 14]. Since image sequences in our dataset are already aligned, the burst-imaging pipeline we compare to is idealized: it benefits from perfect alignment, which is not present in practice. Since alignment is already taken care of, we perform burst denoising by taking the per-pixel median for a sequence of 8 images.

Comparison in terms of PSNR/SSIM using the reference long-exposure images would not be fair to BM3D and burst processing, since these baselines have to use input images that undergo different processing. For fair comparison, we reduce color bias by using the white balance coefficients of the reference image. In addition, we scale the images given to the baselines channel-by-channel to the same mean values as the reference image. These adjustments bring the images produced by the baselines closer in appearance to the reference image in terms of color and brightness. Note that this amounts to using privileged information to help the baselines.

To evaluate the relative quality of images produced by our pipeline, BM3D denoising, and burst denoising, we conduct a perceptual experiment based on blind randomized

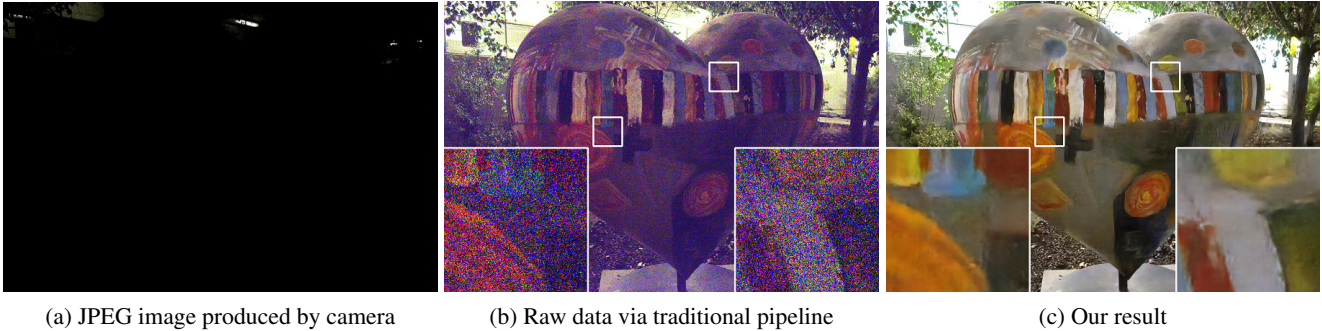

(a) JPEG image produced by camera  (b) Raw data via traditional pipeline  (c) Our result

Figure 5. (a) An image captured at night by the Fujifilm X-T2 camera with ISO 800, aperture f/7.1, and exposure of 1/30 second. The illuminance at the camera is approximately 1 lux. (b) Processing the raw data by a traditional pipeline does not effectively handle the noise and color bias in the data. (c) Our result obtained from the same raw data.

A/B tests deployed on the Amazon Mechanical Turk platform [4]. Each comparison presents corresponding images produced by two different pipelines to an MTurk worker, who has to determine which image has higher quality. Image pairs are presented in random order, with random left-right order, and no indication of the provenance of different images. A total of 1180 comparisons were performed by 10 MTurk workers. Table 2 shows the rates at which workers chose an image produced by the presented pipeline over a corresponding image produced by one of the baselines. We performed the experiment with images from two subsets of the test set: Sony x300 (challenging) and Sony x100 (easier). Our pipeline significantly outperforms the baselines on the challenging x300 set and is on par on the easier x100 set. Recall that the experiment is skewed in favor of the baselines due to the oracle preprocessing of the data provided to the baselines. Note also that burst denoising uses information from 8 images with perfect alignment.

|  | Sony x300 set | Sony x100 set |
| --- | --- | --- |
| Ours > BM3D | 92.4% | 59.3% |
| Ours > Burst | 85.2% | 47.3% |

Table 2. Perceptual experiments were used to compare the presented pipeline with BM3D and burst denoising. The experiment is skewed in favor of the baselines, as described in the text. The presented single-image pipeline still significantly outperforms the baselines on the challenging x300 set and is on par on the easier x100 set.

**Qualitative results on smartphone images.** We expect that best results will be obtained when a dedicated network is trained for a specific camera sensor. However, our preliminary experiments with cross-sensor generalization indicate that this may not always be necessary. We have applied a model trained on the Sony subset of SID to images captured by an iPhone 6s smartphone, which also has a Bayer filter array and 14-bit raw data. We used an app to manually set

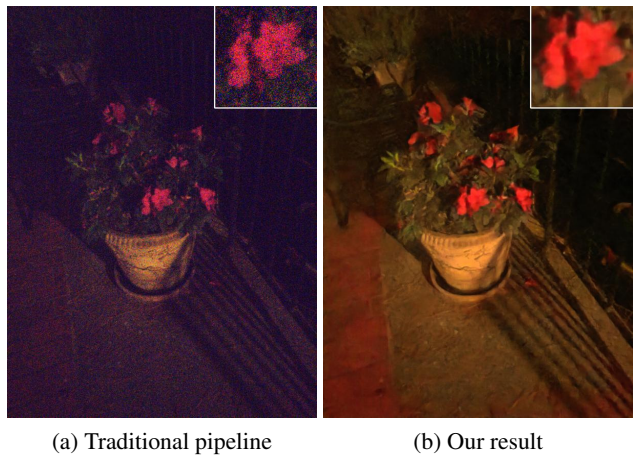

(a) Traditional pipeline  (b) Our result

Figure 6. Application of a network trained on SID to a low-light raw image taken with an iPhone 6s smartphone. (a) A raw image captured at night with an iPhone 6s with ISO 400, aperture f/2.2, and exposure time 0.05s. This image was processed by the traditional image processing pipeline and scaled to match the brightness of the reference image. (b) The output of our network, with amplification ratio x100.

ISO and other parameters, and exported raw data for processing. A representative result is shown in Figure 6. The low-light data processed by the traditional pipeline suffers from severe noise and color shift. The result of our network, trained on images from a different camera, has good contrast, low noise, and well-adjusted color.

### 5.2. Controlled experiments

Table 3 (first row) reports the accuracy of the presented pipeline in terms of Peak Signal-to-Noise Ratio (PSNR) and Structural SIMilarity (SSIM) [38]. We now describe a sequence of controlled experiments that evaluate the effect of different elements in the pipeline.

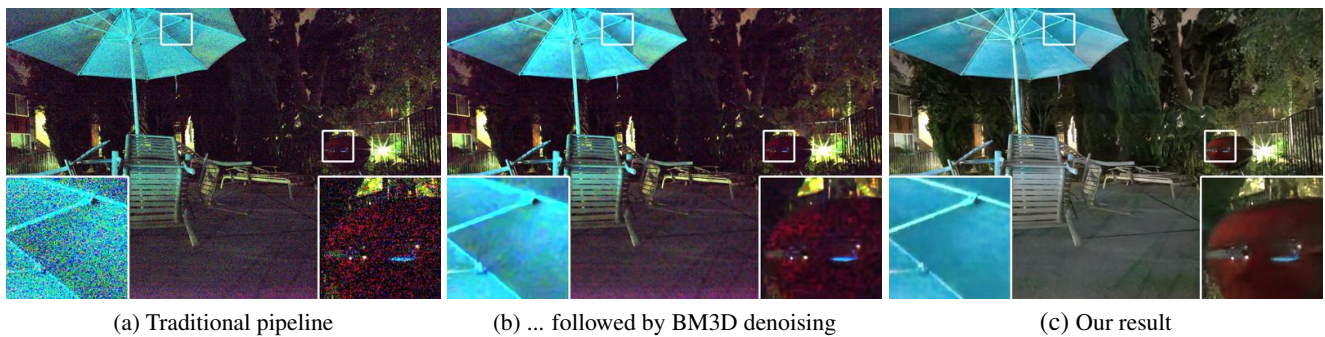

(a) Traditional pipeline     (b) ... followed by BM3D denoising     (c) Our result

Figure 7. An image from the Sony x300 set. (a) Low-light input processed by the traditional image processing pipeline and linear scaling. (b) Same, followed by BM3D denoising. (c) Our result.

| Condition | Sony | Fuji |
|---|---|---|
| 1. **Our default pipeline** | 28.88/0.787 | 26.61/0.680 |
| 2. U-net → CAN | 27.40/0.792 | 25.71/0.710 |
| 3. Raw → sRGB | 17.40/0.554 | 25.11/0.648 |
| 4. $L_1$ → SSIM loss | 28.64/0.817 | 26.20/0.685 |
| 5. $L_1$ → $L_2$ loss | 28.47/0.784 | 26.51/0.680 |
| 6. Packed → Masked | 26.95/0.744 | – |
| 7. X-Trans $3 \times 3$ → $6 \times 6$ | – | 23.05/0.567 |
| 8. Stretched references | 18.23/0.674 | 16.85/0.535 |

Table 3. Controlled experiments. This table reports mean PSNR/SSIM in each condition.

**Network structure.** We begin by comparing different network architectures. Table 3 (row 2) reports the result of replacing the U-net [35] (our default architecture) by the CAN [5]. The U-net has higher PSNR on both sets. Although images produced by the CAN have higher SSIM, they sometimes suffer from loss of color. A patch from the Fuji x300 set is shown in Figure 8. Here colors are not recovered correctly by the CAN.

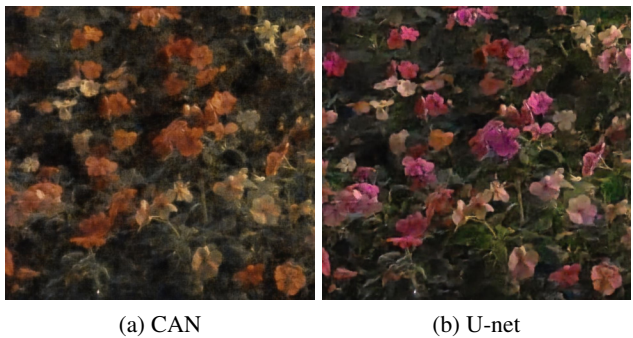

(a) CAN     (b) U-net

Figure 8. Comparison of network architectures on an image patch from the Fuji x300 test set. (a) Using the CAN structure, the color is not recovered correctly. (b) Using the U-net. Zoom in for detail.

**Input color space.** Most existing denoising methods operate on sRGB images that have already been processed by a traditional image processing pipeline. We have found that operating directly on raw sensor data is much more effective in extreme low-light conditions. Table 3 (row 3) shows the results of the presented pipeline when it's applied to sRGB images produced by the traditional pipeline.

**Loss functions.** We use the $L_1$ loss by default, but have evaluated many alternative loss functions. As shown in Table 3 (rows 4 and 5), replacing the $L_1$ loss by $L_2$ or SSIM [43] produces comparable results. We have not observed systematic perceptual benefits for any one of these loss functions. Adding a total variation loss does not improve accuracy. Adding a GAN loss [11] significantly reduces accuracy.

**Data arrangement.** The raw sensor data has all colors in a single channel. Common choices for arranging raw data for a convolutional network are packing the color values into different channels with correspondingly lower spatial resolution, or duplicating and masking different colors [10]. We use packing by default. As shown in Table 3 (row 6), masking the Bayer data (Sony subset) yields lower PSNR/SSIM than packing; a typical perceptual artifact of the masking approach is loss of some hues in the output.

The X-Trans data is very different in structure from the Bayer data and is arranged in $6 \times 6$ blocks. One option is to pack it into 36 channels. Instead, we exchange some values between neighboring elements to create a $3 \times 3$ pattern, which is packed into 9 channels. As shown in Table 3 (row 7), $6 \times 6$ packing yields lower PSNR/SSIM; a typical perceptual artifact is loss of color and detail.

**Postprocessing.** In initial experiments, we included histogram stretching in the processing pipeline for the reference images. Thus the network had to learn histogram stretching in addition to the rest of the processing pipeline. Despite trying many network architectures and loss functions, we were not successful in training networks to per-form this task. As shown in Table 3 (row 8), the accuracy of the network drops significantly when histogram stretching is applied to the reference images (and thus the network has to learn histogram stretching). Our experiments suggest that our pipeline does not easily learn to model and manipulate global histogram statistics across the entire image, and is prone to overfitting the training data when faced with this task. We thus exclude histogram stretching from the pipeline and optionally apply it as postprocessing. Figure 9 shows a typical result in which attempting to learn histogram stretching yields visible artifacts at test time. The result of training on unstretched reference images is darker but cleaner.

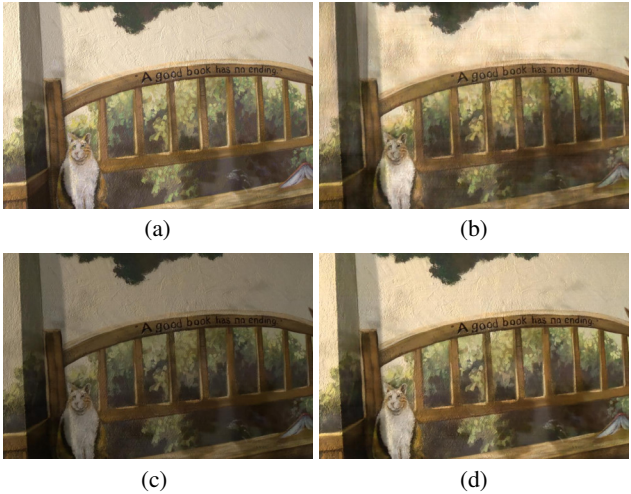

Figure 9. Effect of histogram stretching. (a) A reference image in the Sony x100 set, produced with histogram stretching. (b) Output if trained on histogram-stretched images. The result suffers from artifacts on the wall. (c) Output if trained on images without histogram stretching. The result is darker but cleaner. (d) The image (c) after histogram stretching applied in postprocessing.

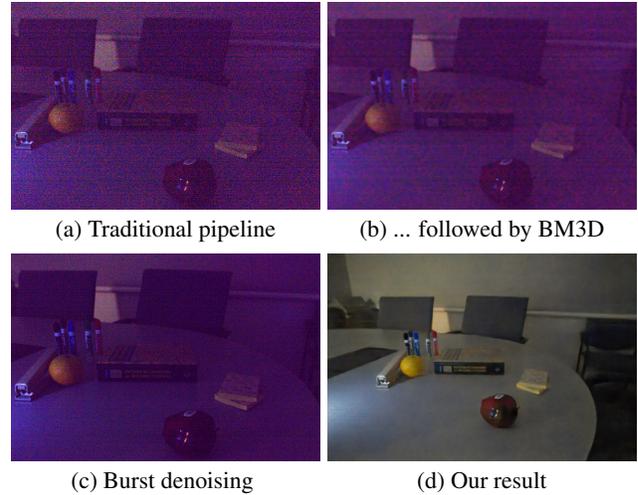

Figure 10. Limited signal recovery in extreme low-light conditions (indoor, dark room, 0.2 lux). (a) An input image in the Sony x300 set, processed by the traditional pipeline and amplified to match the reference. (b) BM3D denoising applied to (a). (c) Burst denoising with 8 images: the result is still bad due to the severe artifacts in all images in the burst. (d) The result of our network; loss of detail is apparent upon close examination.

## 6. Discussion

Fast low-light imaging is a formidable challenge due to low photon counts and low SNR. Imaging in the dark, at video rates, in sub-lux conditions, is considered impractical with traditional signal processing techniques. In this paper, we presented the See-in-the-Dark (SID) dataset, created to support the development of data-driven approaches that may enable such extreme imaging. Using SID, we have developed a simple pipeline that improves upon traditional processing of low-light images. The presented pipeline is based on end-to-end training of a fully-convolutional network. Experiments demonstrate promising results, with successful noise suppression and correct color transformation on SID data.

The presented work opens many opportunities for future research. Our work did not address HDR tone mapping. (Note the saturated regions in Figure 1(c).) The SID dataset is limited in that it does not contain humans and dynamic objects. The results of the presented pipeline are imperfect and can be improved in future work; the x300 subset is particularly challenging. Some artifacts in the output of the presented approach are demonstrated in Figure 10(d).

Another limitation of the presented pipeline is that the amplification ratio must be chosen externally. It would be useful to infer a good amplification ratio from the input, akin to Auto ISO. Furthermore, we currently assume that a dedicated network is trained for a given camera sensor. Our preliminary experiments with cross-sensor generalization are encouraging, and future work could further study the generalization abilities of low-light imaging networks.

Another opportunity for future work is runtime optimization. The presented pipeline takes 0.38 and 0.66 seconds to process full-resolution Sony and Fuji images, respectively; this is not fast enough for real-time processing at full resolution, although a low-resolution preview can be produced in real time.

We expect future work to yield further improvements in image quality, for example by systematically optimizing the network architecture and training procedure. We hope that the SID dataset and our experimental findings can stimulate and support such systematic investigation.

# References

[1] F. Agostinelli, M. R. Anderson, and H. Lee. Adaptive multi-column deep neural networks with application to robust image denoising. In *NIPS*, 2013. 2

[2] J. Anaya and A. Barbu. RENOIR – A dataset for real low-light image noise reduction. *arXiv:1409.8230*, 2014. 2

[3] H. C. Burger, C. J. Schuler, and S. Harmeling. Image denoising: Can plain neural networks compete with BM3D? In *CVPR*, 2012. 2

[4] Q. Chen and V. Koltun. Photographic image synthesis with cascaded refinement networks. In *ICCV*, 2017. 6

[5] Q. Chen, J. Xu, and V. Koltun. Fast image processing with fully-convolutional networks. In *ICCV*, 2017. 4, 7

[6] Y. Chen and T. Pock. Trainable nonlinear reaction diffusion: A flexible framework for fast and effective image restoration. *IEEE Transactions on Pattern Analysis and Machine Intelligence*, 39(6), 2017. 2

[7] K. Dabov, A. Foi, V. Katkovnik, and K. Egiazarian. Image denoising by sparse 3-D transform-domain collaborative filtering. *IEEE Transactions on Image Processing*, 16(8), 2007. 2, 5

[8] X. Dong, G. Wang, Y. Pang, W. Li, J. Wen, W. Meng, and Y. Lu. Fast efficient algorithm for enhancement of low lighting video. In *IEEE International Conference on Multimedia and Expo*, 2011. 2

[9] M. Elad and M. Aharon. Image denoising via sparse and redundant representations over learned dictionaries. *IEEE Transactions on Image Processing*, 15(12), 2006. 2

[10] M. Gharbi, G. Chaurasia, S. Paris, and F. Durand. Deep joint demosaicking and denoising. *ACM Transactions on Graphics*, 35(6), 2016. 2, 7

[11] I. Goodfellow, J. Pouget-Abadie, M. Mirza, B. Xu, D. Warde-Farley, S. Ozair, A. Courville, and Y. Bengio. Generative adversarial nets. In *NIPS*, 2014. 7

[12] S. Gu, L. Zhang, W. Zuo, and X. Feng. Weighted nuclear norm minimization with application to image denoising. In *CVPR*, 2014. 2

[13] X. Guo, Y. Li, and H. Ling. LIME: Low-light image enhancement via illumination map estimation. *IEEE Transactions on Image Processing*, 26(2), 2017. 2

[14] S. W. Hasinoff, D. Sharlet, R. Geiss, A. Adams, J. T. Barron, F. Kainz, J. Chen, and M. Levoy. Burst photography for high dynamic range and low-light imaging on mobile cameras. *ACM Transactions on Graphics*, 35(6), 2016. 1, 2, 4, 5

[15] K. Hirakawa and T. W. Parks. Joint demosaicing and denoising. *IEEE Transactions on Image Processing*, 15(8), 2006. 2

[16] Z. Hu, S. Cho, J. Wang, and M.-H. Yang. Deblurring low-light images with light streaks. In *CVPR*, 2014. 1

[17] V. Jain and H. S. Seung. Natural image denoising with convolutional networks. In *NIPS*, 2008. 2

[18] H. Jiang, Q. Tian, J. E. Farrell, and B. A. Wandell. Learning the image processing pipeline. *IEEE Transactions on Image Processing*, 26(10), 2017. 4

[19] N. Joshi and M. F. Cohen. Seeing Mt. Rainier: Lucky imaging for multi-image denoising, sharpening, and haze removal. In *ICCP*, 2010. 2

[20] J. Kim, J. K. Lee, and K. M. Lee. Accurate image super-resolution using very deep convolutional networks. In *CVPR*, 2016. 4

[21] D. P. Kingma and J. Ba. Adam: A method for stochastic optimization. In *ICLR*, 2015. 5

[22] Y. LeCun, B. Boser, J. S. Denker, D. Henderson, R. E. Howard, W. Hubbard, and L. D. Jackel. Backpropagation applied to handwritten zip code recognition. *Neural Computation*, 1(4), 1989. 4

[23] C. Liu and W. T. Freeman. A high-quality video denoising algorithm based on reliable motion estimation. In *ECCV*, 2010. 2

[24] Z. Liu, L. Yuan, X. Tang, M. Uyttendaele, and J. Sun. Fast burst images denoising. *ACM Transactions on Graphics*, 33(6), 2014. 1, 2, 5

[25] J. Long, E. Shelhamer, and T. Darrell. Fully convolutional networks for semantic segmentation. In *CVPR*, 2015. 4

[26] K. G. Lore, A. Akintayo, and S. Sarkar. LLNet: A deep autoencoder approach to natural low-light image enhancement. *Pattern Recognition*, 61, 2017. 2, 5

[27] A. Łoza, D. R. Bull, P. R. Hill, and A. M. Achim. Automatic contrast enhancement of low-light images based on local statistics of wavelet coefficients. *Digital Signal Processing*, 23(6), 2013. 2

[28] J. Mairal, F. Bach, J. Ponce, G. Sapiro, and A. Zisserman. Non-local sparse models for image restoration. In *ICCV*, 2009. 2

[29] H. Malm, M. Oskarsson, E. Warrant, P. Clarberg, J. Hasselgren, and C. Lejdfors. Adaptive enhancement and noise reduction in very low light-level video. In *ICCV*, 2007. 2

[30] S. Park, S. Yu, B. Moon, S. Ko, and J. Paik. Low-light image enhancement using variational optimization-based Retinex model. *IEEE Transactions on Consumer Electronics*, 63(2), 2017. 2

[31] G. Petschnigg, R. Szeliski, M. Agrawala, M. Cohen, H. Hoppe, and K. Toyama. Digital photography with flash and no-flash image pairs. *ACM Transactions on Graphics*, 23(3), 2004. 2

[32] T. Plötz and S. Roth. Benchmarking denoising algorithms with real photographs. In *CVPR*, 2017. 1, 2, 5

[33] J. Portilla, V. Strela, M. J. Wainwright, and E. P. Simoncelli. Image denoising using scale mixtures of Gaussians in the wavelet domain. *IEEE Transactions on Image Processing*, 12(11), 2003. 2

[34] T. Remez, O. Litany, R. Giryes, and A. M. Bronstein. Deep convolutional denoising of low-light images. *arXiv:1701.01687*, 2017. 1, 4

[35] O. Ronneberger, P. Fischer, and T. Brox. U-net: Convolutional networks for biomedical image segmentation. In *MICCAI*, 2015. 4, 5, 7

[36] L. I. Rudin, S. Osher, and E. Fatemi. Nonlinear total variation based noise removal algorithms. *Physica D: Nonlinear Phenomena*, 60(1-4), 1992. 2

[37] W. Shi, J. Caballero, F. Huszár, J. Totz, A. P. Aitken, R. Bishop, D. Rueckert, and Z. Wang. Real-time single image and video super-resolution using an efficient sub-pixel convolutional neural network. In *CVPR*, 2016. 4

[38] Z. Wang, A. C. Bovik, H. R. Sheikh, and E. P. Simoncelli. Image quality assessment: From error visibility to structural similarity. *IEEE Transactions on Image Processing*, 13(4), 2004. 6

[39] J. Xie, L. Xu, and E. Chen. Image denoising and inpainting with deep neural networks. In *NIPS*, 2012. 2

[40] L. Xu, J. Ren, Q. Yan, R. Liao, and J. Jia. Deep edge-aware filters. In *ICML*, 2015. 4

[41] K. Zhang, W. Zuo, Y. Chen, D. Meng, and L. Zhang. Beyond a Gaussian denoiser: Residual learning of deep CNN for image denoising. *IEEE Transactions on Image Processing*, 26(7), 2017. 2, 4

[42] X. Zhang, P. Shen, L. Luo, L. Zhang, and J. Song. Enhancement and noise reduction of very low light level images. In *ICPR*, 2012. 1

[43] H. Zhao, O. Gallo, I. Frosio, and J. Kautz. Loss functions for image restoration with neural networks. *IEEE Transactions on Computational Imaging*, 3(1), 2017. 7